\documentclass[12pt]{article} 

\usepackage{authblk} 
\usepackage{amssymb}
\usepackage{array}
\usepackage{soul}
\usepackage[normalem]{ulem}
\usepackage{cancel}
\usepackage{amsmath}
\usepackage{hyperref}
\usepackage{url}
\usepackage{lineno}

\usepackage{xcolor}
\usepackage{colortbl}
\usepackage{graphicx}
\usepackage{array}

\begin{document}




\title{GBDTSVM: Combined  Support Vector Machine and Gradient Boosting Decision Tree Framework for efficient snoRNA-disease association prediction} 

\author[1]{Ummay Maria Muna}
\author[1]{Fahim Hafiz}
\author[1]{Shanta Biswas}
\author[1]{Riasat Azim\thanks{Corresponding author: \href{mailto:riasat@cse.uiu.ac.bd}{riasat@cse.uiu.ac.bd}}}

\affil[1]{Department of Computer Science and Engineering, United International University,\\ United City, Madani Avenue, Badda, Dhaka 1212, Bangladesh}

\date{} 
\maketitle 

\begin{abstract}
Small nucleolar RNAs (snoRNAs) are increasingly recognized for their critical role in the pathogenesis and characterization of various human diseases. Consequently, the precise identification of snoRNA-disease associations (SDAs) is essential for the progression of diseases and the advancement of treatment strategies. However, conventional biological experimental approaches are costly, time-consuming, and resource-intensive; therefore,  machine learning-based computational methods offer a promising solution to mitigate these limitations. This paper proposes a model called 'GBDTSVM', representing a novel and efficient machine learning approach for predicting snoRNA-disease associations 
by leveraging a Gradient Boosting Decision Tree (GBDT) and Support Vector Machine (SVM). 'GBDTSVM' effectively extracts integrated snoRNA-disease feature representations utilizing GBDT and SVM is subsequently utilized to classify and identify potential associations. Furthermore, the method enhances the accuracy of these predictions by incorporating Gaussian kernel profile similarity for both snoRNAs and diseases. Experimental evaluation of the GBDTSVM model demonstrated superior performance compared to state-of-the-art methods in the field, achieving an area under the receiver operating characteristic (AUROC) of 0.96 and an area under the precision-recall curve (AUPRC) of 0.95 on MDRF dataset. Moreover, our model shows superior performance on two more datasets named LSGT and PsnoD. Additionally, a case study on the predicted snoRNA-disease associations verified the top 10 predicted snoRNAs across nine prevalent diseases, further validating the efficacy of the GBDTSVM approach. These results underscore the model's potential as a robust tool for advancing snoRNA-related disease research. Source codes and datasets our proposed framework can be obtained from: \url{https://github.com/mariamuna04/gbdtsvm}.

\noindent \textbf{Published Version:} \href{https://doi.org/10.1016/j.compbiomed.2025.110219}{https://doi.org/10.1016/j.compbiomed.2025.110219}
\end{abstract}









\section{Introduction}

With the recent advancement of sequencing technologies and the development of novel methodologies, numerous studies have identified key associations between snoRNAs and human diseases \cite{okugawa2017clinical} \cite{cui2021nop10} \cite{krishnan2016profiling} \cite{chauhan2024snoRNAs} \cite{chabronova2024cardiovascular}. Recent studies have demonstrated the critical role of snoRNA in important biological processes that contribute to the development and progression of conditions such as cancer and hereditary disorders and often work as potential therapeutic targets or diagnostic tools \cite{wajahat2021emerging} \cite{bratkovic2019functional} \cite{chauhan2024snoRNAs} \cite{zacchini2024cancer} \cite{chabronova2024cardiovascular} \cite{shen2024snoRNA}. snoRNAs are a type of non-coding RNA with approximately 60–300 nucleotides, usually found in the nucleoli of eukaryotic cells. 
In 2002, the association of imbalanced snoRNA with cancer was initially proven by down-regulating h5sn2 expressed in normal human brain tissue \cite{wajahat2021emerging} \cite{zhang2023emerging}. Moreover, Chen et al. predicted 120 different snoRNAs and 135 2’-O-ribose methylation sites in rice rRNAs \cite{chen2003high}. 943 snoRNAs have been annotated in the human genome since then \cite{antonarakis2016content}. Along with their identification, research on snoRNA has also revealed their participation in various physiological and pathological cellular processes \cite{wajahat2021emerging}. Okugawa et al. identified SNORA42 as a possible biomarker for colorectal cancer, while Krishnan et al. found a large number of snoRNAs that were associated with a favorable prognosis in breast cancer
\cite{okugawa2017clinical} \cite{krishnan2016profiling}. Furthermore, snoRNAs can be ideal candidates for cancer due to their active participation in tumorigenesis, tumor aggressiveness, and staging. Another study showed that SNHG2 suppresses tumor growth in lung cancer, but its downregulation influences cancer progression \cite{lin2020non}. Since, snoRNAs play such a comprehensive role,  effective investigation of the snoRNAs-disease association (SDA) has gained attention for discovering various complex biomarkers along with disease diagnosis and prognosis in recent times. Traditional computational methods for discovering such associations are costly, time-consuming, and require extensive manual curating; thus, resulting in low identification of the intricate relationships between snoRNAs and diseases.

Numerous computational methods have been proposed so far to predict the associations of diseases with snoRNAs as well as other different types of non-coding RNAs such as long ncRNAs, small ncRNAs, microRNAs, circular RNAs \cite{liu2022identification} \cite{ji2021aemda} \cite{turgut2024dcda} \cite{yan2022pdmda}  \cite{shi2022heterogeneous}. Zhang et al. developed a computational method 'iSnoDi-LSGT' to identify snoRNA-disease association leveraged snoRNA sequences and disease similarity as local similarity constraints \cite{zhang2022isnodi}. Their topological similarities were computed as global topological constraints and used to identify snoRNA-disease associations. Sun et al. proposed a ranking framework called 'iSnoDi-MDRF' that used a matrix completion approach, unlike computational predictors that use few known associations \cite{zhang2023isnodi}. Using the 5-fold stratified shuffling; this approach achieved an AUC of 0.90 and an area under the precision-recall curve (AUPRC) of 0.55. Furthermore, Liu et al. presented GCNSDA, a novel graph neural network (GNN) based model to identify the snoRNA-disease association by using the bipartite graph of snoRNAs and diseases \cite{liu2021gcnsda}. Such a method uncovered latent factors that lead to the underlying connection and provided a guided embedded representation to identify the SDAs. This approach achieved an average AUC of 0.8865 using the advanced GNN algorithm and 5-fold cross-validation.
Zhang et al. utilized self-supervised contrastive learning followed by a light graph convolutional network (LightGCN) named 'GCLSDA' \cite{zhang2023graph}. Contrastive learning was incorporated to deal with sparsity and over-smoothing inside correlation matrices. 'GCLSDA' also included random noise into the generated SnoRNA embedding to increase the true positive rate and finally acquired average AUC and AUPR of 0.9148 and 0.9354, respectively. 'SAGESDA' by Biffon et al. also incorporated GNN along with the attention mechanism to generate new node embeddings using local neighbor nodes, and then the subgraphs were derived from the original graph using the gradient descent technique for better prediction purposes \cite{momanyi2024sagesda}. This framework achieved an AUC value of 0.92 which is comparatively better than the previous GNN-based approach. Generally, in snoRNA-disease association tasks, studies have to deal with the imbalance nature of the dataset. Different studies dealt with imbalanced datasets in different ways; for example, Barrera et al. deals with imbalanced datasets by generating synthetic images for underrepresented classes using a GAN and applying data augmentation techniques to other minority classes \cite{barrera2024deeplearning}. \\
Although recent methods could effectively predict and verify snoRNA disease associations, many of them utilized deep neural networks, making it a black box model lacking interpretability.
\noindent

To address the limitations identified in previous studies, we present GBDTSVM, a novel snoRNA-disease associations prediction method that uses a gradient-boosting decision tree followed by support vector machine (SVM) algorithm to derive SDAs using a known snoRNA-disease association, disease semantic similarity, snoRNA features, and snoRNA functional similarity. The performance of this model is evaluated by 5-fold cross-validation and the experimental results show better performance with an average AUROC score of 0.96 and an average AUPR score of 0.95 , outperforming the other state-of-the-arts (SOTA) methods. We experimented primarily on MDRF dataset, later, we further evaluated our model on two more distinct datasets to demonstrate its scalability and effectiveness. Case studies were conducted to determine the effectiveness of our proposed method. In these case studies, for 12 different malignant diseases, the top predicted snoRNAs were collected and verified from the RNADisease database v4.0 \cite{chen2023rnadisease} and PubMed\cite{liu2019lnc}. Additionally, many known associations were removed and stored as inference data, and the model was further trained and validated on the rest of the datasets. GBDTSVM could successfully predict all excluded associations in the data balancing stage, correctly showing a stable and higher accuracy compared to the SOTA methods. Our primary goal in this method is to identify the roles of snoRNAs in diseases. Therefore, we  propose a novel method to identify the associations of different snoRNAs with different diseases.

\noindent
\section{Materials and Methods}
\subsection{Dataset}
In this study, the primary dataset was obtained from iSnoDi-MDRF, which contains experimentally proven known associations \cite{zhang2023isnodi}. The dataset consists of 111 diseases, and 334 snoRNAs \cite{zhang2023isnodi}. The original source of this dataset is the Mammalian ncRNA-Disease Repository (MNDR) v3.0 database to construct datasets \cite{ning2021mndr}. After eliminating redundant data, they filtered 1010 experimentally supported and predicted snoRNA-disease associations. The dataset contains the features of the snoRNAs that is mainly originated from Durinck et al.\cite{durinck2009mapping}. Moreover, the disease semantic similarity is collected from Wang et al. \cite{wang2007new}. Furthermore, we have experimented on two more datasets named iSnoDi-LSGT that contains 571 snoRNAs and 60 diseases; and PsnoD with 220 snoRNAs and 27 diseases \cite{zhang2022isnodi} \cite{sun2022psnod}.

\noindent
\subsection{Method}
This section discusses the details of GBDTSVM which consists of two steps: data preparation (from section 2.2.1 to 2.2.9), and classifier method (2.2.10). We first balance the dataset, and then the balanced dataset works as an input into the GBDT to extract feature vectors from the combined snoRNA-disease pairs. Lastly, SVM utilizes the derived feature vectors for classifying the snoRNA-disease pairs. Although GBDT could be used directly as a classifier,  this combined approach addresses potential overfitting issues and exploits the properties of the transformed feature space to potentially achieve better performance.

\begin{figure}
    \centering
    \includegraphics[width=1\linewidth]{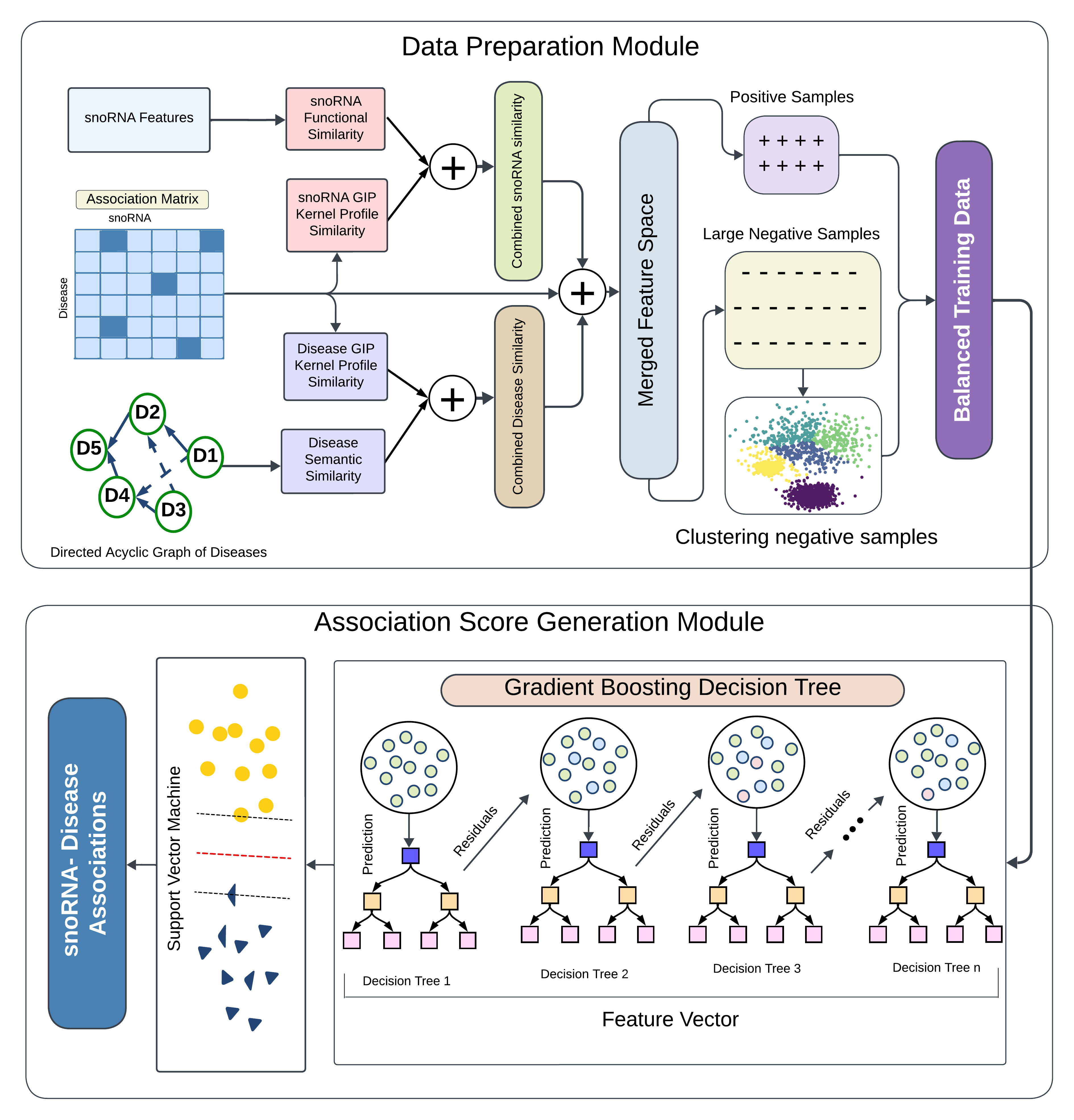}
    \caption{Initialy snoRNA features are utilized to derive snoRNA functional similarity by applying cosine similarity. Meanwhile, disease semantic similarities are measured from the disease DAG graph. Then gaussian kernel profile similarities were measured from the Association matrix for both of the similarity profiles and concatenated and divided into positive and negative samples that were the inputs of the GBDT model after being balanced by applying the clustering mechanism. Finally, the feature vectors generated from the GBDT model are trained using the SVM model and classified the snoRNA-disease association pairs. }
    \label{fig1_method}
\end{figure}

As represented in the diagram in \textbf{Figure 1}, Initially SnoRNA functional similarities were derived from snoRNA features, and disease semantic similarities were generated from a directed acyclic graph of diseases. To enhance the similarity measure, Gaussian profile kernel similarity is applied on both snoRNA and disease features. Subsequently, the corresponding similarities are aggregated for all known and unknown pairs of snoRNA disease. However, the number of positive and negative samples were highly imbalanced, that is why we have incorporated the clustering technique for a large number of negative samples and took the negative samples proportionally equal to the positive ones so that the data became balanced.

\noindent
 \subsubsection{Association Prediction}
     The known associations are a total of 1010 samples, while the unknown associations are \textit{RXD}, which is 37520, where \textit{R} is the total number of snoRNAs and \textit{D} is the total number of diseases. Thus, it is comprehensively visible that the positive and negative samples extensively differ in size; intuitively, the model is prone to be biased towards the negative samples and ignore the positive samples due to this imbalanced nature of classes. To balance the proportion of negative and positive samples, K-Means clustering is applied where the value of \textit{k}  is selected and validated from an experimental perspective. Clustering divides the data into smaller, and more homogeneous groups. By increasing the number of clusters, the algorithm can capture finer-grained details and ensure more diversity in data. Therefore, while training, we experimented with different K (number of clusters) values from 5 to 30, and  a cluster that generated a comparatively higher ROC score and a lower mean squared error(MSE) compared to other K values were selected. Additionally, a higher number of clusters helps to hold variations within the negative samples with more accuracy. In \textbf{Figure \ref{fig1_method}'s} data preparation module, the clustering-based data balancing technique is demonstrated. The corresponding number of negative samples is selected based on the proportion of each group to the total number of unknown samples rather than randomly selecting the same number of negative samples from each group as shown in Figure \textbf{\ref{fig1_method}}. The Gaussian kernel similarity profiles for both snoRNAs and diseases were further included to enhance the similarity measure. The descriptions of these similarity profile measures and known associations are described in the following sections.

\noindent
\subsubsection{Known snoRNA Disease Association}
The dataset consists of both known and unknown associations, with the number of known associations relatively small compared to the dataset. These known associations serve as crucial reference points for the model, acting as markers that guide the extraction of meaningful patterns from the data. Focusing on these positive pairs allows the model to learn the underlying relationships between variables, enabling more accurate inference and generalization across unknown associations. Despite their limited quantity, the associations play a pivotal role in shaping the model's capacity to recognize and predict patterns within the broader dataset.  The associations are denoted as equation 1:


\setcounter{equation}{0}  
\renewcommand{\theequation}{\arabic{equation}}  
\begin{equation}
 D = D^+_{\text{all}} \cup D^-_{\text{all}} 
\end{equation}

where $D^+_{\text{all}}$ and $D^-_{\text{all}}$ are the positive and negative  association of snoRNA-disease, respectively. 

\subsubsection{snoRNA functional similarity}

The functional similarities of the SnoRNAs were derived from the features of the snoRNA. This feature extraction was carried out by Durinck et al., where they extracted DNA and RNA sequences using the refseq IDs of snoRNA and appropriate gene symbols \cite{durinck2009mapping}. These sequences were then used to generate features from the Pse-in-One2.0 web server that uses the pseudo-K-tuple nucleotide composition (PseKNC) method proposed by Chen et al.  \cite{chen2014pseknc}. The dataset comprised 335 snoRNAs, each characterized by 1,025 features, which were selected for further analysis \cite{zhang2023isnodi}. Lastly, we measured the similarity of snoRNA-snoRNA by applying cosine similarity to each of the snoRNA pairs s(i) and s(j) from the feature table and constructed a matrix snoRNA Functional Similarity (SFS) of size 335x335, where SFS (i, j) denotes the functional similarity score between snoRNA s(i) and snoRNA s(j), shown in equation 2.
\begin{equation}
    SFS (i, j) = Cosine\_similarity(s(i) ,  s(j) )
\end{equation}

\subsubsection{ Disease Semantic similarity}
To quantify semantic information and assess disease similarities, the semantic contributions of the ancestor terms for each disease were aggregated, forming a Directed Acyclic Graph (DAG) \cite{wang2007new}. This derived disease semantic similarity has been successfully applied to the other predictors for the identification of the RNA-disease association. Equation 3 denotes the process of calculating the semantic similarities of the diseases using the DAG.
The similarity between two diseases \(d_1\) and \(d_2\) is represented as \(DSS(d_1, d_2)\):

\begin{equation}
DSS(d_1, d_2) = \frac{\sum_{d_i \in A_{d_1} \cap A_{d_2}} (SC_{d_1}(d_i) + SC_{d_2}(d_i))}{\sum_{d_j \in A_{d_1}} SC_{d_1}(d_j) + \sum_{d_j \in A_{d_2}} SC_{d_2}(d_j)}
\end{equation}

\noindent
Where \(DSS(d_1, d_2)\) denotes the Semantic Similarity of the disease \(d_1\) and \(d_2\). \(A_{d_1}\) and \(A_{d_2}\) are the sets of disease nodes containing all ancestors of the diseases \(d_1\) and \(d_2\), respectively, including diseases as nodes in DAG. \(SC_{d_1}(d_i)\) and \(SC_{d_2}(d_i)\) are the semantic contribution values of disease \(d_i\) to diseases \(d_1\) and \(d_2\), respectively.
\noindent
The calculation of the semantic contribution value \(SC_{d_x}(d_i)\) is shown in equation 4:
\begin{equation}
\begin{cases} 
SC_{d_x}(d_i) = \max \{\Delta \cdot SC_{d_x}(d_j) \mid d_j \in \text{children of } d_i\} & \text{if } d_i \neq d_x \\
S_{d_x}(d_i) = 1 & \text{otherwise}
\end{cases}
\end{equation}

\noindent
\(SC_{d_x}(d_i)\) is the semantic contribution value of the disease \(d_i\) to a disease \(d_x\). \(\Delta\) is the decay factor of the semantic contribution, which was set to 0.5 to allow an ancestor disease of a specific disease to have a meaningful impact on its semantics \cite{wang2007new}. It is evident from equation 4 that the contribution of disease \(d_x\) to its own semantic value is 1, whereas the contribution of its ancestor diseases to the semantic value of disease \(d_x\) which is denoted by \(SC_{d_x}(d_i)\), is determined by the maximum semantic contribution of its child, multiplied by \(\Delta\) if \(d_i\) is not equal to \(d_x\), which ensures that only the most significant  contribution of the child node is considered.

\subsubsection{Similarity enhancement of snoRNAs and diseases using Gaussian Kernel Profile similarity}
We further incorporated a gaussian interaction profile kernel to capture complex and nonlinear relationships and to identify the most probable new associations within the interaction profile by leveraging limited known association data for both snoRNAs and diseases. This GIP kernel trick has been widely used in many drug-disease or RNA-disease association prediction tasks. For example, Chen et al. \cite{chen2016wbsmda} and Zhou et al. \cite{zhou2020predicting} included the GIP kernel similarity for miRNA-disease association prediction. On a similar note, Lu et al. \cite{lu2018dr2di} and Jiang et al. \cite{jiang2019predicting} calculated disease similarity using the GIP kernel for the drug-disease association prediction. Using this GIP kernel, researchers predict that diseases with similar association profiles are likely to have similar molecular associations; for example, different subtypes of lung cancer often bind to the same drug molecule. This technique complements the lack of proper interaction data, thus helping the model to predict a better association pattern with shared interaction patterns of diseases or snoRNAs. Therefore, integrating GIP kernel similarity eventually will enhance the representation of the underlying similarities of the disease and snoRNA metrics.

\subsubsection{Gaussian Interaction Profile Kernel Disease Similarity}
To further enhance the disease semantic similarity matrix, we incorporated GIP kernel similarity to calculate the similarity between two diseases, \(d(i)\) and \(d(j)\) using the binary column vectors of the association profile of snoRNA and the disease symbolized by AP, indicating the presence or absence of interactions between entities where rows represent the snoRNAs and columns represent diseases. Equation 5 was utilized to calculate GIP(Gaussian interaction profile) Kernel similarity between the association profiles and to capture how similar the entities are in terms of their interaction patterns. The equation is as follows:
\begin{equation}
GIP_{\text{dis}}(d(i), d(j)) = \exp(-\partial_d \| AP(d(i)) - AP(d(j)) \|^2),
\end{equation}
\(GIP_{\text{dis}}\) denotes the similarity between disease profiles, which is calculated by the squared Euclidean distance of the \(i_{\text{th}}\) and \(j_{\text{th}}\) column vectors, then multiplied with an adjustable parameter \(\partial_d\) for smoothness and to normalize the kernel bandwidth that ensures the appropriate similarity measurement. In addition, \(\partial_d\) in equation 6 is calculated using the sum of the squared column vector of the diseases, which is divided by the total number of diseases, and lastly multiplied by the \(\partial_d\)’, whose value is set to 1  \cite{van2011gaussian} :

\begin{equation}
\partial_d = \partial'_d \left( \frac{1}{n_d} \sum_{i=1}^{n_d} \| AP(d(i)) \|^2 \right).
\end{equation}

\subsubsection{Gaussian Interaction Profile Kernel snoRNA Similarity}
GIP kernel similarity for snoRNAs is also calculated using the same formula used for the disease similarity, however, in this case, the binary vectors are taken from the rows of association profile(AP) since the row represents the snoRNAs. Thus the similarity between two snoRNA, s(i) and s(j) is calculated using equation 7:
\begin{equation}
GIP_{\text{sno}}(s(i), s(j)) = \exp(-\partial_s \| AP(s(i)) - AP(s(j)) \|^2),
\end{equation}
\noindent
Here, \(GIP_{\text{sno}}\) represents the similarities between each snoRNA pair which is calculated using binary row vectors of \(AP\) consisting of a particular snoRNA's association with all the diseases. The squared distance of the \(i_{\text{th}}\) and \(j_{\text{th}}\) rows are multiplied by the kernel adjusting parameter \(\partial_s\)’ that is calculated using equation 8 for all rows of snoRNAs.
\begin{equation}
\partial_s = \partial'_s \left( \frac{1}{n_s} \sum_{i=1}^{n_s} \| AP(s(i)) \|^2 \right).
\end{equation}

\subsubsection{Meshed Similarity for snoRNAs and diseases}
The snoRNA functional similarity matrix may lack similarities for some pairs, but GIP kernel profile similarities are available for all pairs based on their association profiles. To address this, we merged the two profiles into a unified similarity profile for snoRNAs, which is called Meshed snoRNA Functional Similarity (MSFS), as shown in Figure \textbf{\ref{fig1_method}}. The calculation for MSFS is given in equation 9. Similarly, the GIP kernel profile similarity for diseases was combined with the disease semantic similarity, creating the Meshed Disease Semantic Similarity (MDSS), and the calculation is similar to MSFS. MSFS and MDSS profiles provide a more comprehensive and robust similarity measure for predictive modeling. Pairs of both similarity profiles were integrated by averaging them.

\begin{equation}
MSFS(s(i), s(j)) = 
\begin{cases} 
\frac{GIP_{\text{sno}}(s(i), s(j)) + SFS(s(i), s(j))}{2}; & \begin{array}{l} \text{if } m(i) \text{ and } m(j) \text{ have} \\ \text{functional similarity} \end{array} \\
GIP_{\text{sno}}(s(i), s(j)); & \text{otherwise}
\end{cases}
\end{equation}



\subsubsection{Feature Extraction using Gradient Boosting Decision Tree}

Using balanced positive and negative pairs with their similarity measures, new feature vectors are generated. The Gradient Boosting Decision Tree works in employed for this task as shown in Figure \textbf{\ref{fig1_method}}. This technique builds an ensemble of decision trees and operates iteratively. It updates the weights of each tree based on the negative gradients of the loss function from previous classifiers. Through grid search strategy, we selected the optimal parameters of GBDT using 5-fold cross-validation (CV), where the optimal tree number is 10. The main task of the GBDT works with initializing the first weak classifier \(DT_0(x)\) utilizing the optimal loss function, as shown in equation 10. Since no DT has been trained yet, labels are initially set to an average value, denoted by avg. The equation is as follows: 
\begin{equation}
DT_0(x) = \arg\min_c \sum_{i=1}^N L(y_i, avg)
\end{equation}
\noindent
Based on the loss generated from the \(DT_0\), Argmin tries to minimize the loss function. The residuals are calculated using Equation 11 and are updated for the next classifier:

\begin{equation}
r_{ci} = -\left[\frac{\partial L(y_i, DT(x_i))}{\partial DT(x_i)}\right]_{DT(x) = DT_{c-1 }(x)}
\end{equation}
\noindent
Where \(r_{ci}\) is the residual value of the \(i_{th}\) sample of classifier c, whose value is the negative derivation of the loss function concerning the previous prediction by \(DT(x_i)\).
\noindent
Next, the regression tree is trained using the features \(x_i\) against \(r_i\) and resulting in the creation of the leaf node \(L_{ck}\) where \(k\) indicates the leaves of the classifier \(c\). The objective is to find a \(\gamma_{ck}\) that minimizes the loss function at the \(k_{th}\) leaf node, as defined by Equation 12.

\begin{equation}
\gamma_{ck} = \arg\min_\gamma \sum_{i \in L_{ck}} L(y_i, DT_{c-1}(x_i) + \gamma)
\end{equation}
\noindent
The weight \(\gamma\) is computed by aggregating the loss across all the samples \(x_i\) that belong to the leaf node \(L_{ck}\). Subsequently, the model is updated by adding the derived \(\gamma_{ck}\) to the previous prediction \(DT_{c-1}(x)\), resulting in the updated prediction \(DT_c(x)\), as shown in equation 13. 

\begin{equation}
DT_c(x) = DT_{c-1}(x) + \sum_{k=1}^K \gamma_{ck} \mathbb{I}_{L_{ck}}(x)
\end{equation}
\noindent
where the function \(\mathbb{I}_{L_{ck}}(x)\) used to denote whether a given input \(x\) falls into a specific leaf node \(L_{ck}\) in the \(c_{th}\) iteration, assign value 1 if its argument is true, and 0 otherwise. The previous steps are repeated 10 times, which is equal to the total number of decision trees in GBDT model that finally makes a stronger classifier \(DT_C(x)\) by iteratively adding the contributions of multiple decision trees where each tree correct the errors of the previous trees. Additional explanations of these equations have been added to \textbf{supplementary document 2}. Now, using the log-likelihood function \(L( DT(x_i), y_i)\), the loss is calculated in each iteration. The new feature vector is constructed using the outputs of the decision trees of the gradient boosting model. In each iteration, the GBDT model trains and updates its base classifiers and makes a new decision learning from the previous one. The decision trees split the data based on the concatenated snoRNA features and disease features, aiming to minimize the loss function and end up with multiple leaf nodes. Each input feature \(x_i\) falls on a specific leaf node for each regression tree, Thus, the value for that leaf node is set to 1 if a sample falls into it, otherwise, it is set to 0. A new feature vector is created for each sample where each element corresponds to a leaf node from all the decision trees, and the combined vector length equals the total number of leaf nodes across all the regression trees, hence the vectors consist of 0s and 1s. This derived features from GBDT enhance classification performance by capturing complex interactions within biological data. Then, by mapping samples to leaf nodes and encoding them via one-hot transformation, GBDT identifies non-linear patterns that may reflect underlying biological associations. Finally, the vector consisting of the features was trained using 5 fold CV method that divided the data into five parts, with four parts are used for training and the remaining one for testing.

\subsubsection{Final Classification by Support Vector Machine}


Passing feature vectors from the GBDT model into an SVM leads the SVM to apply a Radial Basis Function kernel, transforming the data into a higher-dimensional space. This transformation allows the SVM to separate data that were not originally separable. The main target of SVM is to find the best hyperplane to separate the data into two classes so that the margin between the classes is maximized while performing correct classification. Finally, whenever the model selects the optimal hyperplane, it classifies the feature vectors into two classes to determine whether snoRNA is associated or not with disease. 
A Support Vector Machine with an RBF kernel was configured for the classification tasks. To optimize the model’s performance, a parameter grid was defined to adjust the hyperparameters, including different values for the regularization parameter and the kernel coefficients. The grid search conducts an in-depth search over the specified parameter grid with 5-fold cross-validation to find the optimal hyperparameters for the SVM model. While incorporating SVM, we follow the soft margin optimization strategy is shown in Equation 14, which allows some missclassifications by introducing slack variables, enabling a trade-off between maximizing the margin and minimizing classification errors.
\begin{equation}
   \begin{aligned}
    & \min_{\mathbf{w}, b} \quad \frac{1}{2} \|\mathbf{w}\|^2 + C \sum_{i=1}^{n} \xi_i \\
    & \text{subject to} \quad y_i (\mathbf{w} \cdot \mathbf{x}_i + b) \geq 1 - \xi_i, \quad \xi_i \geq 0, \quad i = 1, \ldots, n
\end{aligned}
\end{equation}

where \( C \) represents the regularization parameter, w denotes the weight vector, \( b \) is the bias term, \( \ \xi_i\) are the slack variable that allows certain erroneous classifications, \( x_i \) are the input feature vectors, \( y_i \) are the labels, and \( n \) is the total training samples of the SVM model. It intends to minimize \(\frac{1}{2} \|\mathbf{w}\|^2\) to help in finding the hyperplane with the maximum margin between the two classes. Meanwhile, the parameter \( C \) controls the trade-off between achieving a large separable margin and minimizing the classification error.  In this strategy, all feature vectors are classified with a certain confidence score.  

\subsection{Evaluation Measures}
To evaluate the effectiveness of the GBDTSVM model, we used two metrics AUROC (Area Under the ROC Curve) and AUPRC (Area Under the Precision-Recall Curve) which are widely used metrics in binary classification tasks.\\
AUROC score demonstrates the model's ability to distinguish between positive and negative associations. It is measured using the ROC curve, which plots the True Association Rate(TAR) against the False Association Rate(FAR). Equations 18 and 19 show the calculation of TAR and FAR.
\begin{equation}
\text{TAR} = \frac{\text{True Associations}}{\text{True Associations} + \text{False Disassociations}}
\end{equation}
\begin{equation}
\text{FAR} = \frac{\text{False Associations}}{\text{False Associations} + \text{True Disassociations}}
\end{equation}

\noindent
However, AUPRC focuses on precision and recall, making it useful in imbalanced datasets. It measures the trade-off between Precision and Recall. Equations 20 and 21 show the calculation of Precision and Recall.
\begin{equation}
\text{Precision} = \frac{\text{True Associations}}{\text{True Associations} + \text{False Associations}}
\end{equation}
\begin{equation}
\text{Recall} = \frac{\text{True Associations}}{\text{True Associations} + \text{False Disassociations}}
\end{equation}

\noindent
Together, these metrics give a more accurate evaluation of the model's performance in identifying snoRNA-disease associations.

\section{Result and Discussion}
Evidence from extensive research suggest that snoRNAs play pivotal roles in diverse biological processes and can contribute to many molecular mechanisms of pathogens of various diseases. The following sections aim to elucidate the associations between specific snoRNAs and various diseases through integrative computational approaches and evidence from the existing literature.\\

\subsection{Performance Evaluation}
The performance of GBDTSVM was evaluated by testing the data of unknown association pairs based on 5-fold cross-validation. Primarily, we measure the area under the receiver operating characteristics(ROC-AUC) that shows good performance when the value is high. In addition, we calculated the area under the precision recall curve(AUPRC) which shows the trade-off between the model's precision and recall.
\begin{figure}[hbt!]
    \centering
    \includegraphics[width=1\linewidth]{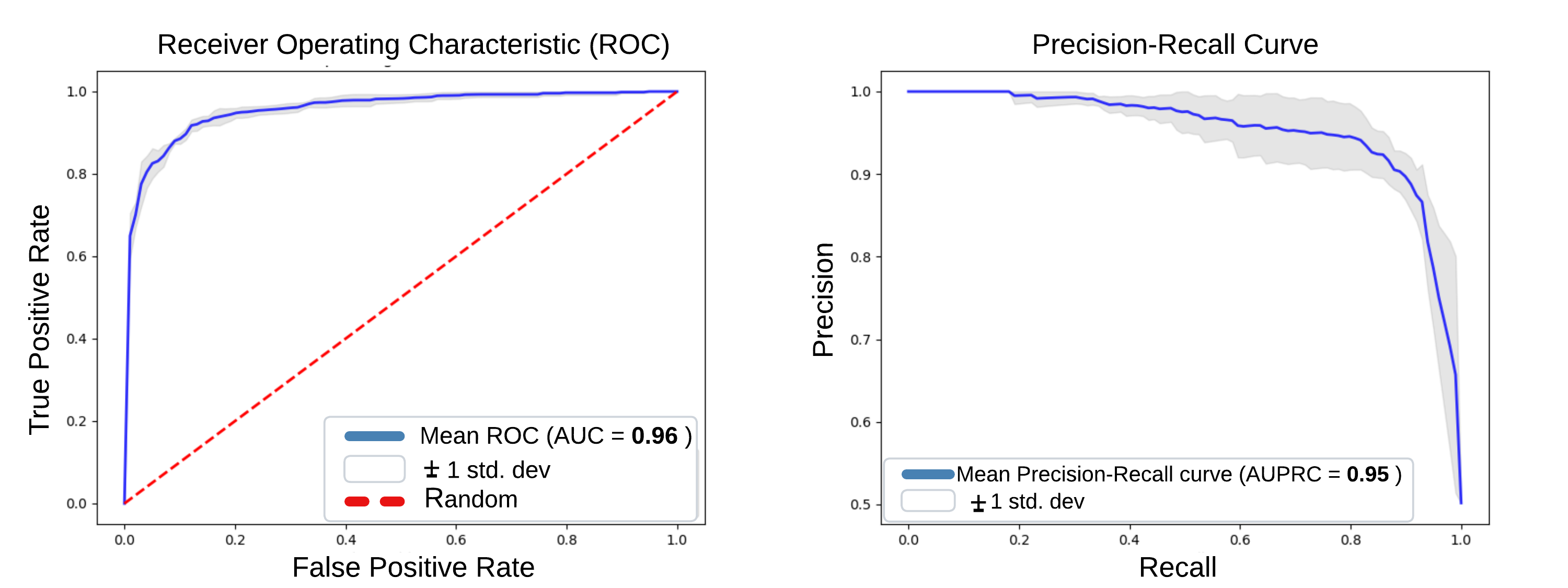}
    \caption{The ROC and P-R Curves of GBDTSVM for snoRNA and disease association identification on MDRF dataset.}
    \label{fig:The ROC and P-R Curves of GBDTSVM for snoRNA and disease association identification}
\end{figure}

GBDTSVM achieved a mean ROC-AUC of 0.96 and an AUPRC of 0.95 on the 5-fold CV as shown in Figure 2 (left and right), respectively. In the 5-fold cross-validation, all known snoRNA-disease associations were randomly divided into five parts, one of which was used as a testing sample while the other four parts were used as training samples. Moreover, the accuracy and F1 score of the model are 0.95 and 0.94, respectively, which also proves the robustness of our method. To demonstrate the scalability of the GBDTSVM model, we have also shown performance in three datasets in terms of all the aforementioned metrics in \textbf{Table 1}. 

\begin{table}[ht!]
\centering
\caption{Performance Metrics of GBDTSVM for the three Datasets} 
\vspace{6pt} 
\begin{tabular}{>{\centering\arraybackslash}m{4cm}>{\centering\arraybackslash}m{2cm}>{\centering\arraybackslash}m{2cm}>{\centering\arraybackslash}m{2cm}}
\hline
\textbf{Metric} & \textbf{MDRF\cite{zhang2023isnodi}} & \textbf{LSGT\cite{zhang2022isnodi}} & \textbf{PsnoD\cite{sun2022psnod}} \\ [4pt]
\hline
ROC-AUC  & 0.97  & 0.93  & 0.95  \\[3pt]
AUPRC & 0.96 & 0.93 & 0.94 \\[3pt]
Accuracy & 0.95  & 0.89  & 0.93  \\[3pt]
F1 Score & 0.94  & 0.85  & 0.92  \\[3pt]
\hline
\end{tabular}
\end{table}


A comparison is performed with some state-of-the-art methods to show the better performance of GBDTSVM compared to the other seven existing methods. We have assessed the prediction capability of GBDTSVM and the other state-of-the-art models based on the scores of ROC-AUC and AUPRC. \textbf{Figure 4} demonstrates that the GBDTSVM outperformed the other methods, such as SAGESDA \cite{momanyi2024sagesda} , GCNSDA \cite{liu2021gcnsda}, GCLSDA \cite{zhang2023graph}, GCNMDA \cite{long2020predicting}, IGCNSDA \cite{hu2024igcnsda}, iSnoDi-MDRF\cite{zhang2023isnodi} and iSnoDi-LSGT\cite{zhang2022isnodi}.
These SOTA models achieved ROC-AUCs of 0.92, 0.88, 0.91, 0.87, 0.84, 0.92, and 0.95; and AUPRs of 0.90, 0.89, 0.93, 0.88, 0.87, 0.37 and 0.75, respectively, as shown in \textbf{Table 2} whereas the GBDTSVM achieved scores of 0.96 and 0.95, as mentioned earlier, that demonstrate its effectiveness in accurately predicting snoRNA-disease associations.\\
Previous works utilized graph neural networks for such snoRNA association prediction, while our method includes proportional clustering techniques to balance the highly imbalanced dataset along with an ensemble model GBDT as feature transformation and SVM for classification. Such a combination is comparatively less resource-intensive as it does not use heavy deep learning models like others and also results in high accuracy compared to the recent work. In addition, our model demonstrates variations in execution time, CPU usage, and memory consumption based on dataset size, efficiently handling both small and large datasets with relatively lower computational costs for smaller datasets. The MDRF dataset takes the highest execution time of 94.48 seconds for 112X335 snoRNA-diseases, PsnoD takes the lowest execution time of 28.39 seconds for 27X220 snoRNA-diseases, and LSGT in between. Other resource  utilizations for all the datasets are given in \textbf{supplementary document 2}. The variations suggest that our model scales differently depending on the size and complexity of the data set.

\begin{figure}[hbt!]
    \centering
    \includegraphics[width=1\linewidth]{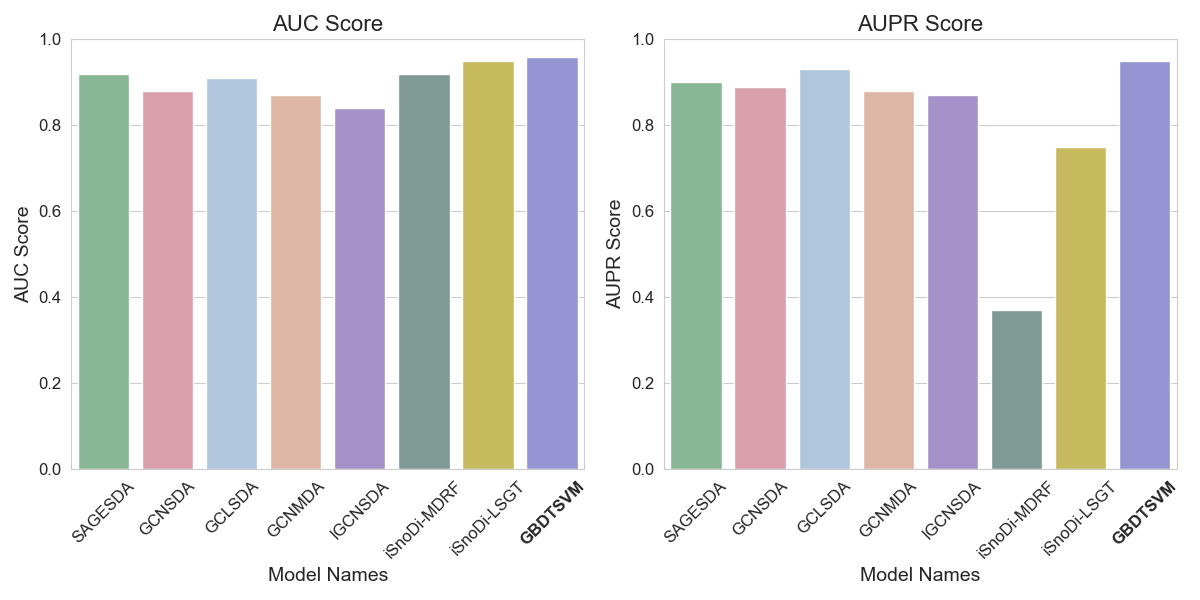}
    \caption{Performance Comparison with SOTA methods in 5-CV showing that GBDTSVM outperforms in terms of AUC and AUPR score across all the methods.}
    \label{fig:Performance Comparison with other methods in 5-CV}
\end{figure}

\begin{table}[ht!]
\centering
\caption{Performance Comparison of GBDTSVM against state-of-the-arts methods}
\begin{tabular}{>{\centering\arraybackslash}m{3cm}>{\centering\arraybackslash}m{2cm}>{\centering\arraybackslash}m{2cm}}
\hline \\
\textbf{Model} & \textbf{AUC} & \textbf{AUPR} \\ [4pt]
\hline \\
SAGESDA & 0.92 & 0.90 \\[3pt]

GCNSDA & 0.88 & 0.89 \\[3pt]

GCLSDA & 0.91 & 0.93 \\[3pt]

GCNMDA & 0.87 & 0.88 \\[3pt]

IGCNSDA & 0.84 & 0.87 \\[3pt]

iSnoDi-MDRF & 0.92 & 0.37\\[3pt]

iSnoDi-LSGT & 0.95 & 0.75 \\[3pt]

\textbf{GBDTSVM}  & \textbf{0.96} & \textbf{0.95} \\[3pt]
\hline
\end{tabular}
\end{table}


\subsection{Case Study}

To illustrate the capability and effectiveness of GBDTSVM, several case studies have been conducted on different prevalent diseases.  As mentioned above, among 37520 associations obtained from the iSnoDi-MDRF database, only 1010 associations were known; For training purposes, the same number of unknown associations was used to balance negative associations with positive associations. However, during prediction, scores were generated for all remaining unknown associations after data balancing. Subsequently, the diseases were ranked in descending order based on their predicted association scores with different snoRNAs. The top 10 pairs were selected from the sorted associations for the 9 diseases and the top 5 pairs for 3 diseases. These associations were verified using the RNADisease database, which contains 343,273 experimentally verified data points in 18 RNA categories, 117 species, and 4,090 diseases, and PubMed, a comprehensive resource for biomedical research. The verified top associations are presented in \textbf{Table 3(top 10 verified snoRNAs for colorectal cancer, osteosarcoma and lung adenocarcinoma)}, \textbf{Table 4(top 10 verified snoRNAs for  ovarian cancer, glioma and primary breast cancer)}, \textbf{Table 5(top 10 verified snoRNAs for gastric cancer, stomach cancer and prostate cancer)}, and \textbf{Table 6(top 5 verified snoRNAs for cirrhosis, Osteoarthritis, and Alzheimer)}. The detailed results are provided in the \textbf{supplementary document 1}.

\begin{table}[ht!]
\centering
\caption{snoRNAs associated with  Colorectal cancer, Osteosarcoma and Lung adenocarcinom}
\begin{tabular}{p{2cm} p{2cm} p{2cm} p{2cm} p{2cm} p{2cm} }
\hline
\multicolumn{2}{c}{Colorectal Cancer} & \multicolumn{2}{c}{Osteosarcoma} & \multicolumn{2}{c}{Lung Adenocarcinoma} \\
\hline \\
snoRNAs & Evidence & snoRNAs & Evidence & snoRNAs & Evidence \\[3pt]
\hline \\
SNHG22 & RNADisease & SNHG8 & RNADisease & SNORA48 & RNADisease \\[2pt]
SNHG15 & RNADisease & SNHG14 & RNADisease & SNORA24 & RNADisease \\[2pt]
SNHG8 & RNADisease & SNHG15 & RNADisease & SNORA70C & RNADisease \\[2pt]
SNORA24 & RNADisease & SNHG16 & RNADisease & SNHG5 & RNADisease \\[2pt]
SNHG16 & RNADisease & SNHG4 & RNADisease & SNHG6 & RNADisease \\[2pt]
SNHG20 & RNADisease & SNHG7 & RNADisease & SNHG7 & RNADisease \\[2pt]
SCARNA9 & RNADisease & SNHG5 & RNADisease & SNHG1 & unconfirmed \\[2pt]
SNHG6 & RNADisease & SNHG6 & RNADisease & SNHG14 & RNADisease \\[2pt]
SNORD33 & RNADisease & SNORD94 & RNADisease & SNHG15 & RNADisease \\[3pt]
SNHG7 & RNADisease & SCARNA9 & RNADisease & SNHG17 & RNADisease \\[5pt]

\hline
\end{tabular}
\end{table}

\begin{table}[ht!]
\centering
\caption{ snoRNAs associated with Ovarian cancer, Glioma and Primary breast cancer}
\begin{tabular}{p{2cm} p{2cm} p{2cm} p{2cm} p{2cm} p{2cm} }
\hline 
\multicolumn{2}{c}{Ovarian cancer} & \multicolumn{2}{c}{Glioma} & \multicolumn{2}{c}{Primary breast cancer} \\
\hline \\
snoRNAs & Evidence & snoRNAs & Evidence & snoRNAs & Evidence \\[3pt]
\hline
SNHG16 & RNADisease & SNHG1 & PubMed\cite{liu2019lnc} & SNHG3 & RNADisease \\[2pt]
SNHG3 & RNADisease & SNHG6 & RNADisease & SNHG16 & RNADisease \\[2pt]
SNHG15 & RNADisease & SNHG3 & PubMed\cite{zhang2020lnc} & SNHG20 & PubMed\cite{guan2018lnc} \\[2pt]
SNHG14 & RNADisease & SNHG7 & RNADisease & SNHG6 & RNADisease \\[2pt]
TERC & PubMed\cite{kushner2000antisense} & SNHG16 & RNADisease & SNHG12 & RNADisease \\[2pt]
SNHG20 & RNADisease & SNHG14 & RNADisease & SNHG14 & RNADisease \\[2pt]
SNHG7 & RNADisease & SNHG12 & PubMed\cite{sun2018lnc} & SNHG15 & RNADisease \\[2pt]
SNHG6 & PubMed\cite{wu2019lnc} & SNHG9 & RNADisease & SNHG7 & PubMed\cite{li2020lnc} \\[2pt]
SNHG5 & RNADisease & SNHG5 & RNADisease & SNHG5 & RNADisease \\[3pt]
SNHG17 & RNADisease & SNHG17 & RNADisease & SNHG8 & RNADisease \\[5pt]

\hline
\end{tabular}
\end{table}

\noindent Numerous additional associations were confirmed through both the RNADisease database and PubMed. In addition, here the analysis of the selected literature for specific disease pair associations is discussed. For example, Shan et al. showed an oncogenic role of SNHG7, which is a small nucleolar RNA host gene 7 in colorectal cancer, as a candidate for colorectal cancer \cite{shan2018lncrna} . Similarly, another study examined the function and mechanism of SNHG22 for the same disease \cite{yao2021lncrna}. Two more recent studies have shown the potential role of SNHG16 in colorectal cancer \cite{chen2024lncRNA} \cite{helmy2024snhg16}. Yang et al. detected SNHG16 as a promoter of cell growth in ovarian cancer \cite{yang2018long}. Two recent studies have highlighted the significant contribution of SNHG6 and SNHG17 to ovarian cancer \cite{liang2024snhg17} \cite{lukina2024snhg6}. Likewise, Oliayi et al. found up-regulation of SNHG6 in primary breast cancers, promoting cell cycle progression in breast cancer-derived cell lines \cite{jafari2019snhg6}. Qi et al. discovered that SNHG7 accelerates prostate cancer expansion and cycle progression through cyclin D1 by sponging miR-503 \cite{qi2018long}. Lastly, Wang et al. show that cell proliferation, migration, invasion, and the epithelial-mesenchymal transition process are promoted by SNHG4 by sponging miR-204-5p in gastric cancer \cite{wang2021lncrna}.\\

\begin{table}[ht!]
\centering
\caption{snoRNAs associated with Gastric cancer, Stomach Cancer and Prostate cancer}
\begin{tabular}{p{2cm} p{2cm} p{2cm} p{2cm} p{2cm} p{2cm} }
\hline
\multicolumn{2}{c}{Gastric cancer} & \multicolumn{2}{c}{Stomach Cancer} & \multicolumn{2}{c}{Prostate cancer} \\
\hline \\
snoRNAs & Evidence & snoRNAs & Evidence & snoRNAs & Evidence \\[3pt]
\hline \\
SNORD94 & RNADisease & SNHG20 & RNADisease & SNHG7 & RNADisease \\[2pt]
SNORD3A & RNADisease & SNHG16 & RNADisease & SNHG12 & RNADisease \\[2pt]
SNORD89 & RNADisease & SNHG14 & RNADisease & SNHG16 & PubMed\cite{weng2021lnc} \\[2pt]
SNORD43 & RNADisease & SNHG12 & RNADisease & SNHG14 & RNADisease \\[2pt]
SNORD97 & RNADisease & SNHG7 & RNADisease & SNHG15 & RNADisease \\[2pt]
SNORD104 & RNADisease & SNHG6 & RNADisease & SNHG8 & PubMed\cite{shi2021lnc} \\[2pt]
SNORA54 & unconfirmed & SNHG15 & RNADisease & SNHG4 & RNADisease \\[2pt]
SNORD116-3 & RNADisease & SNHG5 & RNADisease & SNHG17 & RNADisease \\[2pt]
SNORD50A & RNADisease & SNHG8 & RNADisease & SNHG3 & RNADisease \\[3pt]
SNHG4 & RNADisease & SNHG9 & unconfirmed & SCARNA10 & unconfirmed \\[5pt]

\hline
\end{tabular}
\end{table}

\begin{table}[ht!]
\centering
\caption{RNAs associated with  Liver Cirrhosis, Osteoarthritis and Alzheimer's Disease}
\begin{tabular}{p{2cm} p{2cm} p{2cm} p{2cm} p{2cm} p{2cm} }
\hline
\multicolumn{2}{c}{Liver Cirrhosis} & \multicolumn{2}{c}{Osteoarthritis} & \multicolumn{2}{c}{Alzheimer} \\
\hline \\
snoRNAs & Evidence & snoRNAs & Evidence & snoRNAs & Evidence \\[3pt]
\hline \\
SNHG22 & RNADisease & SNHG5 & RNADisease & SNHG14 & RNADisease \\[2pt]
SNHG7 & RNADisease & SNHG15 & RNADisease & SNHG3 & RNADisease \\[2pt]
SNORD37 & RNADisease & SNHG16 & RNADisease & SNORD115 & RNADisease \\[2pt]
SNHG1 & RNADisease & SNHG14 & RNADisease & SNORD116 & RNADisease \\[3pt]
SNHG5 & RNADisease & SNHG1 & RNADisease & SNHG7 & RNADisease \\[5pt]

\hline
\end{tabular}
\end{table}

The above case study supports the initial hypothesis of our proposed model, confirming its effectiveness in identifying associations of snoRNA diseases. Lastly, our comprehensive literature review indicates that data from different yet closely related domains can be interchangeably utilized for association prediction using our method \cite{zhou2020predicting} \cite{lei2019gbdtcda} \cite{duan2021gbdtlrl2d}\cite{zhou2024boosting} \cite{sun2025multigraph}.


\section{Conclusion}

snoRNA plays a vital role in many biological processes. The irregular or depletion expression of snoRNAs is related to various diseases. Although some studies have shown associations of different snoRNAs with different diseases, specifically cancer diseases, a large number of snoRNAs have not yet been discovered or verified with specific targets that are committed to certain cell functions. Therefore, it is crucial to detect and verify the association of a diverse range of snoRNAs with different prevalent diseases. \\
Furthermore, prediction of snoRNA-disease associations is important for real-world applications in biomedical research and drug discovery. Through the discovery of previously unknown associations between snoRNAs and diseases, this work can provide a deeper understanding of disease mechanisms at the molecular level along with the non-coding RNAs and the potential role ncRNA might play in different diseases. Therefore, these insights may help identify novel biomarkers for disease diagnosis and prognosis. GBDTSVM shows a simple yet effective method to identify the associations of different snoRNAs with different diseases, and also the best associations were verified using case studies and various existing literature. Our proposed method first deals with a major imbalance issue in the positive and negative data with known and unknown associations; and effectively incorporates gradient-boosting decision tree for feature extraction and a support vector machine for the final classification. Using data on snoRNA disease, GBDT extracts meaningful features, which the SVM successfully learns to differentiate known and unknown associations with high prediction accuracy. Finally, we demonstrate that our method outperformed state-of-the-art methods with a comparatively higher prediction score. Such a method might also be useful to predict the association of the disease with other non-coding RNAs, for example miRNAs and lncRNAs. Exploratory biological analyses in this research area might reveal new solutions and findings.

\end{document}